\documentclass[11pt,a4paper]{article}
\usepackage[hyperref]{eacl2021}
\usepackage{times}
\usepackage{latexsym}

\usepackage{graphicx}
\usepackage[ruled, linesnumbered]{algorithm2e}
\usepackage{amssymb,mathtools}
\usepackage{xcolor}
\usepackage{comment}
\usepackage{mathtools}
\usepackage[thinlines]{easytable}
\usepackage{babel,blindtext}

\usepackage{diagbox}
\usepackage{multirow}
\usepackage{longtable}
\usepackage{dblfloatfix}
\usepackage{scrextend}
\usepackage{float}
\usepackage{caption}
\usepackage{url}
\usepackage{multicol}
\usepackage{amsfonts}
\usepackage{subfigure}

\usepackage{array}
\newcolumntype{L}[1]{>{\raggedright\let\newline\\\arraybackslash\hspace{0pt}}m{#1}}
\newcolumntype{C}[1]{>{\centering\let\newline\\\arraybackslash\hspace{0pt}}m{#1}}
\newcolumntype{R}[1]{>{\raggedleft\let\newline\\\arraybackslash\hspace{0pt}}m{#1}}

\usepackage{microtype}

\aclfinalcopy 

\title{NLP-CUET@DravidianLangTech-EACL2021: Offensive Language Detection from Multilingual Code-Mixed Text using Transformers}

\author{Omar Sharif{\textdagger}, Eftekhar Hossain{*} and Mohammed Moshiul Hoque{\textdagger}\\
    {\textdagger}Department of Computer Science and Engineering \\
    {*}Department of Electronics and Telecommunication Engineering\\
    Chittagong University of Engineering and Technology, Bangladesh \\
  \texttt{\{omar.sharif, eftekhar.hossain, moshiul\_240\}@cuet.ac.bd}\\

  }

\date{}

\begin{document}
\maketitle
\begin{abstract}
The increasing accessibility of the internet facilitated social media usage and encouraged individuals to express their opinions liberally. Nevertheless, it also creates a place for content polluters to disseminate offensive posts or contents. Most of such offensive posts are written in a cross-lingual manner and can easily evade the online surveillance systems. This paper presents an automated system that can identify offensive text from multilingual code-mixed data. In the task, datasets provided in three languages including Tamil, Malayalam and Kannada code-mixed with English where participants are asked to implement separate models for each language. To accomplish the tasks, we employed two machine learning techniques (LR, SVM), three deep learning (LSTM, LSTM+Attention) techniques and three transformers (m-BERT, Indic-BERT, XLM-R) based methods. Results show that XLM-R outperforms other techniques in Tamil and Malayalam languages while m-BERT achieves the highest score in the Kannada language. The proposed models gained weighted $f_1$ score of $0.76$ (for Tamil), $0.93$ (for Malayalam), and $0.71$ (for Kannada) with a rank of $3^{rd}$, $5^{th}$ and $4^{th}$ respectively. 
\end{abstract}

\section{Introduction}
The exponential increase of offensive contents in social media has become major concern to government organizations and tech companies. It is impossible to identify offensive texts manually from enormous amounts of online contents generated every moment in social media and other online platforms. Therefore, academicians, policymakers and stakeholders are trying to develop robust computational systems to limit the spread of offensive contents using state of the art NLP techniques \citep{trac-2020-trolling,alw-2020-online}. In the last few years, several studies have been conducted regarding offensive languages such as hate speech, aggression, toxicity and abusive language. Although most social media users used their regional languages to carry out communication; however, plenty of resources developed in English, Arabic \citep{mubarak-etal-2017-abusive} and other high-resources languages. Lately, people started using code-mixed texts on social media. Linguistic level understanding of the text, complex morphological structure, and lack of training corpora are crucial barriers to classifying code-mixed data. Moreover, the system trained on monolingual data alone may fail to classify code-mixed data because of the texts linguistic level code-switching complexity. This work aims to address the above mentioned problems by contributing the following:
\begin{itemize}
    \item Prepared transformer-based methods to identify the offensive texts from multilingual (Tamil, Malayalam, Kannada) code-mixed data. 
    \item Perform experiments on the dataset with detail performance and error analysis, thus setting an important baseline to compare in future.
\end{itemize}

\begin{table*}[b!]
\centering
\begin{tabular}{c|ccc|ccc|ccc}
\hline
 & \multicolumn{3}{c}{\textbf{Tamil}}& \multicolumn{3}{c}{\textbf{Malayalam}}& \multicolumn{3}{c}{\textbf{Kannada}}\\
\hline
&\textbf{Train}&\textbf{Valid}&\textbf{Test}&\textbf{Train}&\textbf{Valid}&\textbf{Test}&\textbf{Train}&\textbf{Valid}&\textbf{Test}\\
\hline
NF & 25425   & 3193  & 3190 &  14153  &  1779 & 1765 & 3544 &  426  &  427  \\
OTIO & 454   & 65    &  71  &  -      &  -    &  -   &   123 &  16  &  14  \\          
OTII & 2343  & 307   & 315  &   239   &  24   &  27  &   487 &  66  &  75  \\   
OTIG & 2557  & 295   & 288  &   140   &  13   &  23  &   329 &  45  &  44  \\   
OU   & 2906  & 356   & 368  &   191   &  20   &  29  &   212 &  33  &   33  \\   
NT   & 1454  & 172   & 160  &    -    &  -    & -    &  -    &  -   &   - \\   
NM   &  -    & -     &  -   &   1287  &  163  & 157  &  -    &  -   &  -  \\   
NK   &  -    & -     &  -   &    -    & -     &  -   & 1522  &  191 & 185   \\ 
\hline
\textbf{Total}  &  23962  & 4850   & 2500  & 58500  & 5842  & 1923 & 36009 & 5724  & 2682 \\ 
\hline            
\end{tabular}
\caption{\label{table1}{Class wise distribution of train, validation and test set for each language. The acronym used for classes abbreviated in Section \ref{Data}}.
}
\end{table*}
\section{Related Work}
Offensive social media content might trigger objectionable consequences to its user like mental health problem, and suicide attempts \citep{bonanno2013cyber}. To keep the social media ecosystem, coherent researchers and stakeholders should try to develop computational models to identify and classify offensive contents within a short period. \citet{zampieri2019predicting} develop an offensive language identification dataset using hierarchical annotation schema. Three layers of annotation are used in the dataset: offensive language detection, categorization of offensive language, and offensive language target identification. SemEval task 6 is organized based on this dataset which had opened interesting research directions \citep{zampieri2019semeval}. In early stages, computational models created by using support vector machine, naive Bayes and other traditional machine learning approach \citep{dadvar2013improving}. These models performance is not up to the mark as they could not capture the semantic and contextual information in texts. This problem was mitigated with the arrival of word embeddings and recurrent neural networks \citep{aroyehun2018aggression}. Networks like bidirectional LSTMs and GRUs can hold contextual information from both past and future, creating more robust classification systems \citep{mandl2019overview}. In recent years, transformer-based model such as BERT \cite{sharif2021combating}, XLM-R \cite{ranasinghe2020multilingual} gained more attention to identify and classify offensive texts. These large pre-trained model can classify code-mixed texts of different languages with astonishing accuracy \citep{hande2020kancmd,chakravarthi2020corpus}.

\section{Dataset}
\label{Data}
In order to detect offensive text from social media, task organizers developed a gold standard corpus. As many social media texts are code-mixed, so system trained on monolingual data fails to classify code-mixed data due to the complexity of code-switching at different linguistic levels in the texts. To address this phenomenon, \citet{chakravarthi2020corpus} developed a code-mixed text corpus in Dravidian languages. Corpus has three types of code-mixing texts (Tamil-English, Kannada-English and Malayalam-English). The task aims to implement a system that can successfully detect code-mixed offensive texts. To implement such a system, we utilize the corpus provided by the workshop organizers\footnote{https://dravidianlangtech.github.io/2021/index.html}\citep{dravidianoffensive-eacl}. The corpus consists of the text of three different languages, i.e. Tamil, Kannada and Malayalam. 

System developed with the different number of train, validation and test examples for each language. The assigned task is a multi-class classification problem where the model has to classify a potential text into one of six predefined classes. These classes are Not offensive (NF), Offensive-Targeted-Insult-Other (OTIO), Offensive-Targeted-Insult-Individual (OTII), Offensive-Targeted-Insult-Group (OTIG), not-Tamil (NT)/not-Malayalam (NM)/not-Kannada (NK), and Offensive-Untargetede (OU). Table \ref{table1} shows the number of instances for each class in train, validation and test sets. Datasets are highly imbalanced where a number of instances in NF class is much higher compare to other classes. Before system development, a cleaning process is applied to every dataset. In this phase, unwanted characters, symbols, punctuation, emojis, and numbers are discarded from the texts and thus, a cleaned dataset prepared for each language.    

\section{Methodology}
\begin{figure}[ht!]
\centering
  \includegraphics[width=\linewidth]{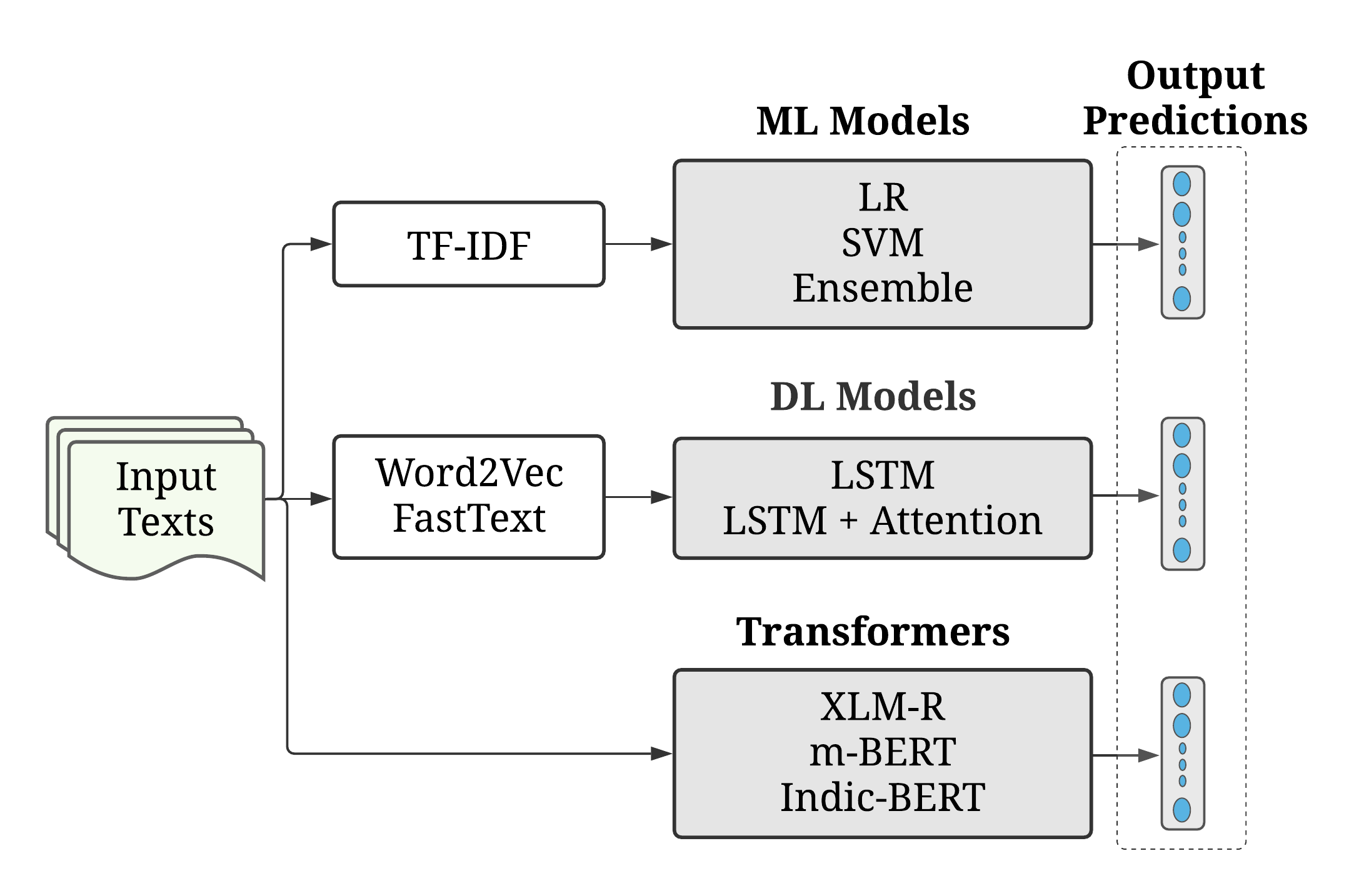}
  \caption{Abstract process of of offensive language detection}
  \label{fig:proposed_model}
\end{figure}

In this part, we briefly describe the methods and techniques employed to address the previous section's problem. First, features are extracted with different feature extraction techniques and various machine learning (ML) and deep learning algorithms are used for the baseline evaluation. Furthermore, different transformer models, i.e. XLM-R, m-BERT and Indic-BERT, are exploited as well. Figure~\ref{fig:proposed_model} shows the schematic process of the developed system.   

\paragraph{Feature Extraction:} ML and DL algorithms are unable to learn from raw texts. Therefore, feature extraction is required to train the classifier models. For ML methods,  tf-idf technique \citep{tokunaga1994text} is applied to extract the unigram features. On the other hand, Word2Vec \citep{mikolov2013distributed} and FastText \citep{grave2018learning} embeddings are used as feature extraction techniques for DL models. Word2Vec implemented using Keras embedding layer with embedding dimension $100$ for all languages while pre-trained embedding matrix of each language is utilized for FastText embedding.   

\paragraph{Machine Learning Models}
To develop the offensive text detection system, we begin our investigation with traditional ML approaches, including LR and SVM. The ensemble of multiple ML classifiers is also employed to achieve better performance. We use `lbfgs' optimizer with $C$ value of $0.4$, $0.7$, and $5$ for Tamil, Malayalam and Kannada languages to implement LR. We selected `linear' SVM and settled $C$ value to $3$, $10$, and $7$. Besides, Decision Tree (DT) and Random Forest (RF) classifiers are incorporated with LR and SVM to construct the ensemble approach. For LR and SVM, the same parameters were retained while `n\_estimators = $100$' is used for DT and RF. Majority voting technique is applied to get the prediction from the ensemble method.       
\paragraph{Deep Learning Models}
DL algorithms are proven superior over many traditional ML approaches in various classification tasks \citep{sharif2020techtexc}. We employ LSTM and combination of LSTM with Attention-based approach to classify the offensive text to continue the investigations. LSTM is well known for its ability to capture the semantic information as well as long term dependencies. Bidirectional LSTM (BiLSTM) with $100$ cells are used to exploit the information from both past and future states. To mitigate the chance of overfitting, a dropout technique utilized with a rate of $0.1$. Finally, the output of the BiLSTM layer transferred to a softmax layer for the prediction. However, sometimes all the words in a text do not contribute equally for the classification. To emphasize the words that have a noticeable impact on the input text, we use attention \citep{vaswani2017attention} mechanism. An attention layer of $20$ neurons is constructed on the top of a BiLSTM layer. The attention operation applied upon the output of BiLSTM layer and the attention vector passed to a softmax layer for the prediction. We employ the same architectures for all the languages and use `sparse\_categorical\_crossentropy’ as the loss function. `Adam’ optimizer with learning rate $1e^{-3}$ and batch size $32$ is used to train the models for 20 epochs. The best intermediate model is stored by using Keras callbacks.
\begin{table*}[b!]
\centering
\begin{tabular}{cl|ccc|ccc|ccc}
\hline
\textbf{Method}&\textbf{Classifiers} & \multicolumn{3}{c}{\textbf{Tamil}}& \multicolumn{3}{c}{\textbf{Malaylam}}& \multicolumn{3}{c}{\textbf{Kannada}}\\
\hline
&&\textbf{P}&\textbf{R}&\textbf{F}&\textbf{P}&\textbf{R}&\textbf{F}&\textbf{P}&\textbf{R}&\textbf{F}\\
\hline
\multirow{3}{*}{ML models}&LR &0.76 & 0.65 & 0.69 & 0.88 & 0.84 &  0.86 &  0.52  & 0.44  &    0.47\\
&SVM &0.74  &0.68 &   0.70 & 0.88 &   0.87 &       0.88  &0.48 &    0.48 & 0.48   \\          
&Ensemble & 0.72 &   0.76 &  0.73 & 0.88 &  0.89&  0.88&0.46  &    0.51 & 0.48 \\                    
\hline            
\multirow{3}{*}{DL models}&LSTM (Word2vec) & 0.73 &0.72 &0.72 & 0.86  & 0.87& 0.86& 0.48 & 0.44&      0.46   \\              
&LSTM (Fasttext) & 0.70 & 0.67  & 0.68 & 0.87&     0.85 &    0.86  & 0.50 &  0.45 &  0.45   \\       
&LSTM + Attention &0.71& 0.73&  0.72& 0.86 & 0.87& 0.87  &0.49  &  0.46  &  0.47     \\          
\hline

\multirow{3}{*}{Transformers}&m-BERT & 0.74 & 0.78& 0.76& 0.93&  0.88 & 0.90   & 0.70 &  0.74&    \textbf{0.71}  \\                     
&Indic-BERT &0.74 &  0.78  &   0.74 &   0.95 & 0.91 &    0.92 & 0.69 & 0.74 & 0.70 \\       
&XLM-R & \textbf{0.75} &  0.78    &  \textbf{0.76} &0.92 &  0.94  &  \textbf{0.93}  & 0.71 & 0.70 & 0.71\\       
\hline
\end{tabular}
\caption{\label{result} Evaluation results of ML, DL and transformer-based models on the test set. Here P, R, F denotes the precision, recall and weighted $f_1$ score.
}
\end{table*}
\paragraph{Transformer Models}
In recent years, transformers have gained popularity for its tremendous performance in almost every aspect of NLP. As our given datasets consist of cross-lingual texts of different languages, we choose three transformers such as XLM-R \citep{conneau2019unsupervised}, m-BERT \citep{devlin2019bert}, Indic-BERT \citep{kakwani2020inlpsuite} to develop our models. XLM-R is a self-supervised training technique for cross-lingual understanding particularly well for low resource languages. On the other hand, m-BERT is a transformer model pre-trained over 104 languages, and Indic-BERT is specifically pre-trained on Indian languages such as Kannada, Tamil, Telugu and Malayalam. These models are culled from Pytorch Huggingface \footnote{ https://huggingface.co/transformers/} transformers library and fine-tuned on our dataset using ktrain \citep{maiya2020ktrain} package. To fine-tune these models, we use ktrain `fit\_onecycle’ method with learning rate $2e^{-5}$ for all the languages. We observed that each input text's average length is less than 50 words for Kannada and 70 words for Tamil and Malayalam languages. Therefore, to reduce the computational cost, the input texts' maximum size settle to 50 for Kannada and 70 for Tamil and Malayalam. All the models have trained up to $20$ epochs with batch size 4 for XLM-R and 12 for m-BERT and Indic-BERT.       

\section{Results and Analysis}
This section presents a performance comparison of various ML, DL, and transformer-based offensive text detection models for all the mentioned languages. The superiority of the models is determined based on the weighted $f_1$ score. However, the precision and recall metrics also considered. Table \ref{result} reports the evaluation results of models. The results showed that the ML models ensemble achieved the highest $f_1$ score of $0.73$ for the Tamil language. Both SVM and ensemble methods obtained a similar $f_1$ score for Malayalam ($0.88$) and Kannada ($0.48$) languages. In contrast, for all the languages, the LR method was poorly performed where a small difference $(<0.04)$ in $f_1$ score value was observed with other ML models. In case of DL techniques, LSTM (with Word2Vec) model obtained comparatively higher $f_1$ score of $0.72$, $0.87$ and $0.47$ than that of provided by LSTM (FastText) modes which are $0.68$, $0.86$, and  $0.45$ for Tamil, Malayalam and Kannada languages. However, a combination of LSTM and Attention model shows $0.01$ rise amounting to $0.87$ and $0.46$ respectively for Malayalam and Kannada languages.  

Meanwhile, transformer-based models showed outstanding performance for all the languages. For Tamil, Indic-BERT obtained $f_1$ score of $0.74$ while both m-BERT and XLM-R exceeds all the previously mentioned models by achieving maximum $f_1$ score of $0.76$. Though two models give an identical $f_1$ score, XLM-R is opted as the best model by considering both the precision, recall scores. In Malayalam language, m-BERT and Indic-BERT obtained $f_1$ score of $0.90$ and $0.92$ respectively. While XLM-R shows a rise of $0.01$ and thus beats all the other models. For the Kannada language, both XLM-R and m-BERT outperformed all the previous models obtaining $f_1$ score of $0.71$. However, m-BERT is selected as the best model as it outdoes XLM-R concerning precision and recall values.    

The results revealed that the transformer-based models performed exceptionally well to accomplish the assigned task for all the languages. Compared to other techniques, the ML models performed moderately while the deep learning models lag than the other models. The possible reason for this weaker performance is the extensive appearance of cross-lingual words in the text. As a result, Word2Vec and FastText embeddings have failed to create the appropriate feature mapping among the words. Thus, LSTM and attention-based models may not find sufficient relational dependencies among the features and performed below the expectation. 

\subsection{Error Analysis}
\begin{figure*}
\begin{multicols}{3}
    \subfigure[Tamil]{\includegraphics[height=4cm, width=0.33\textwidth]{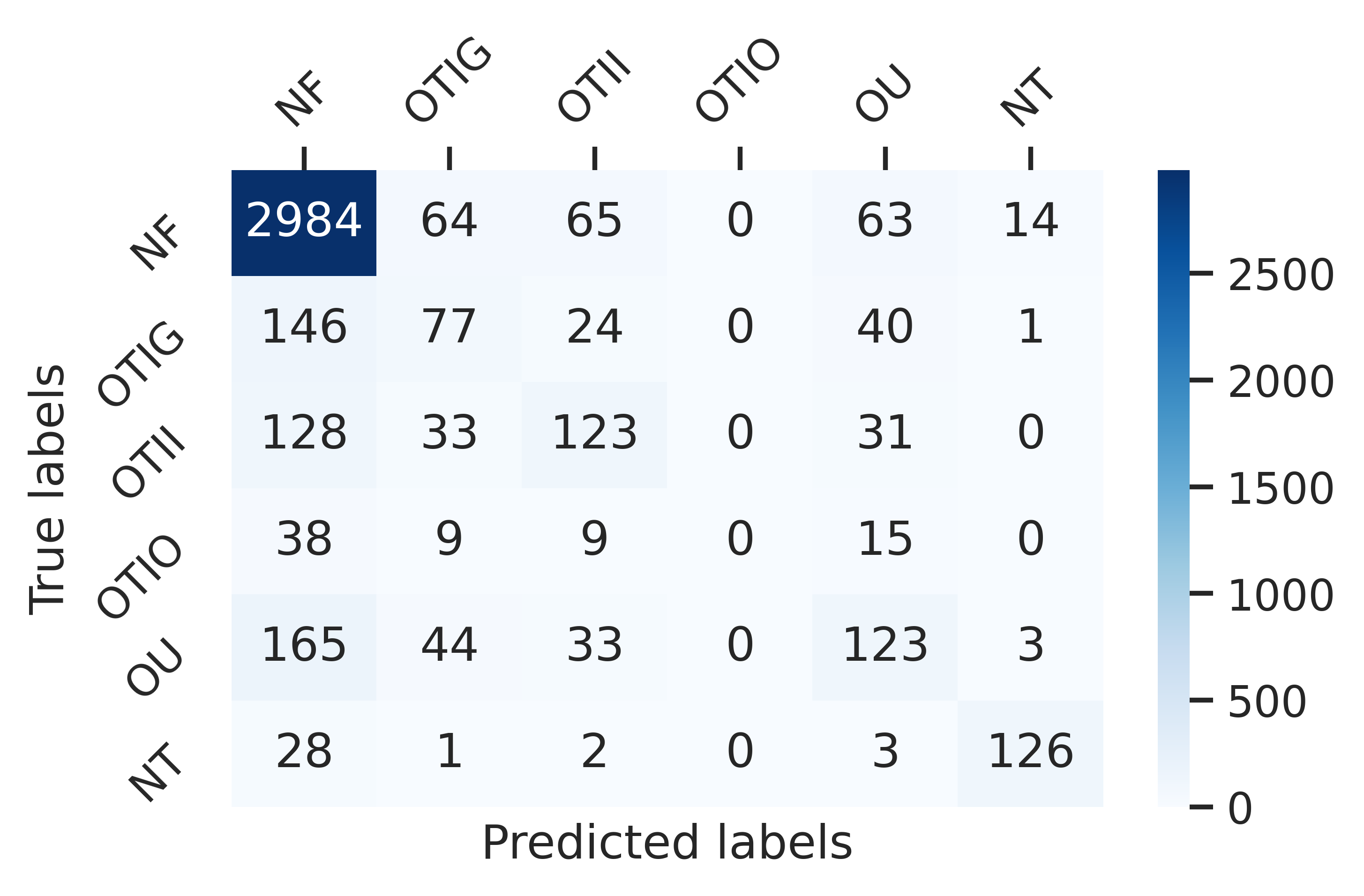}}
    \subfigure[Malayalam]{\includegraphics[height=4cm, width=0.33\textwidth]{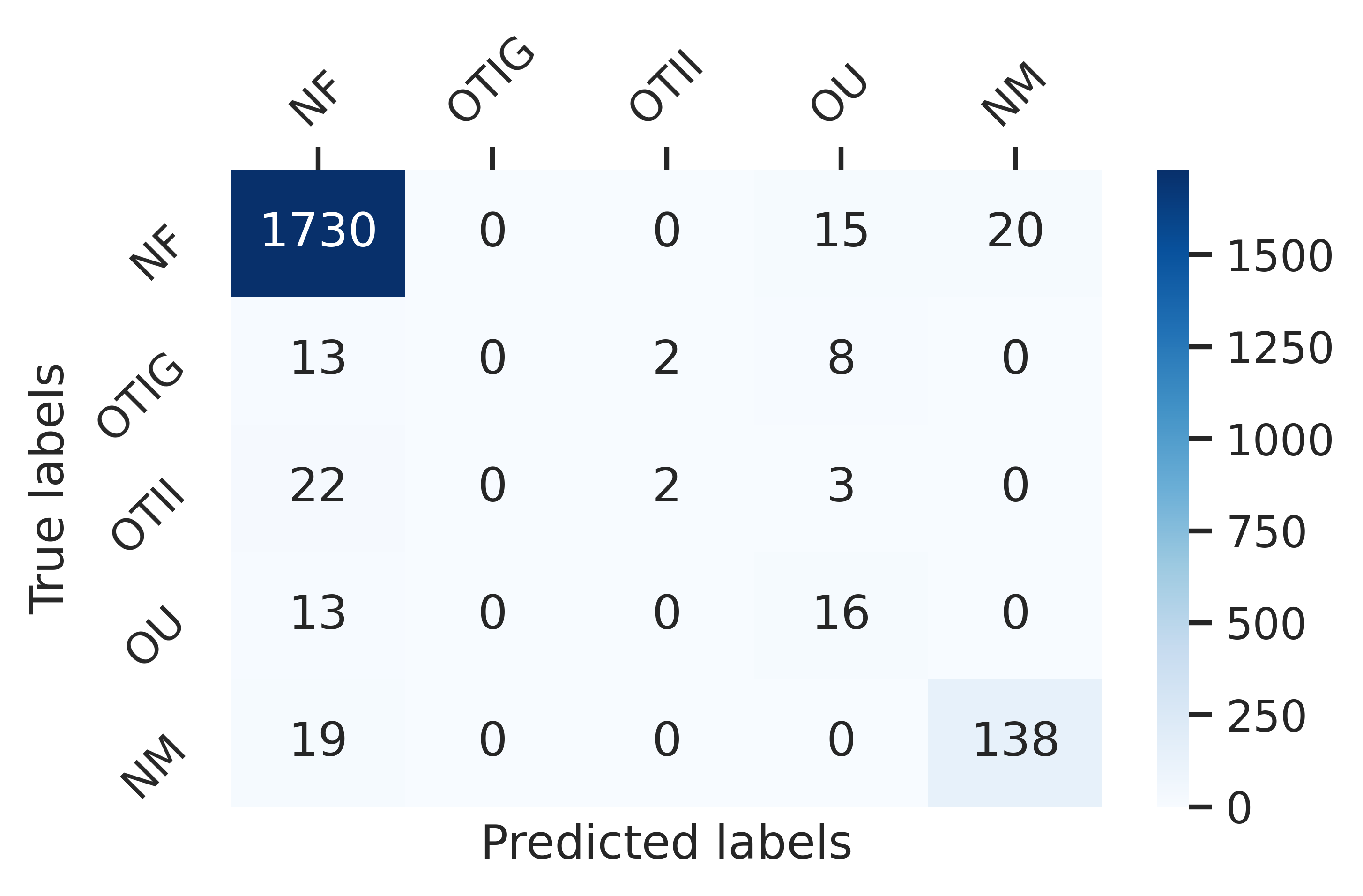}}
    \subfigure[Kannada]{\includegraphics[height=4cm, width=0.33\textwidth]{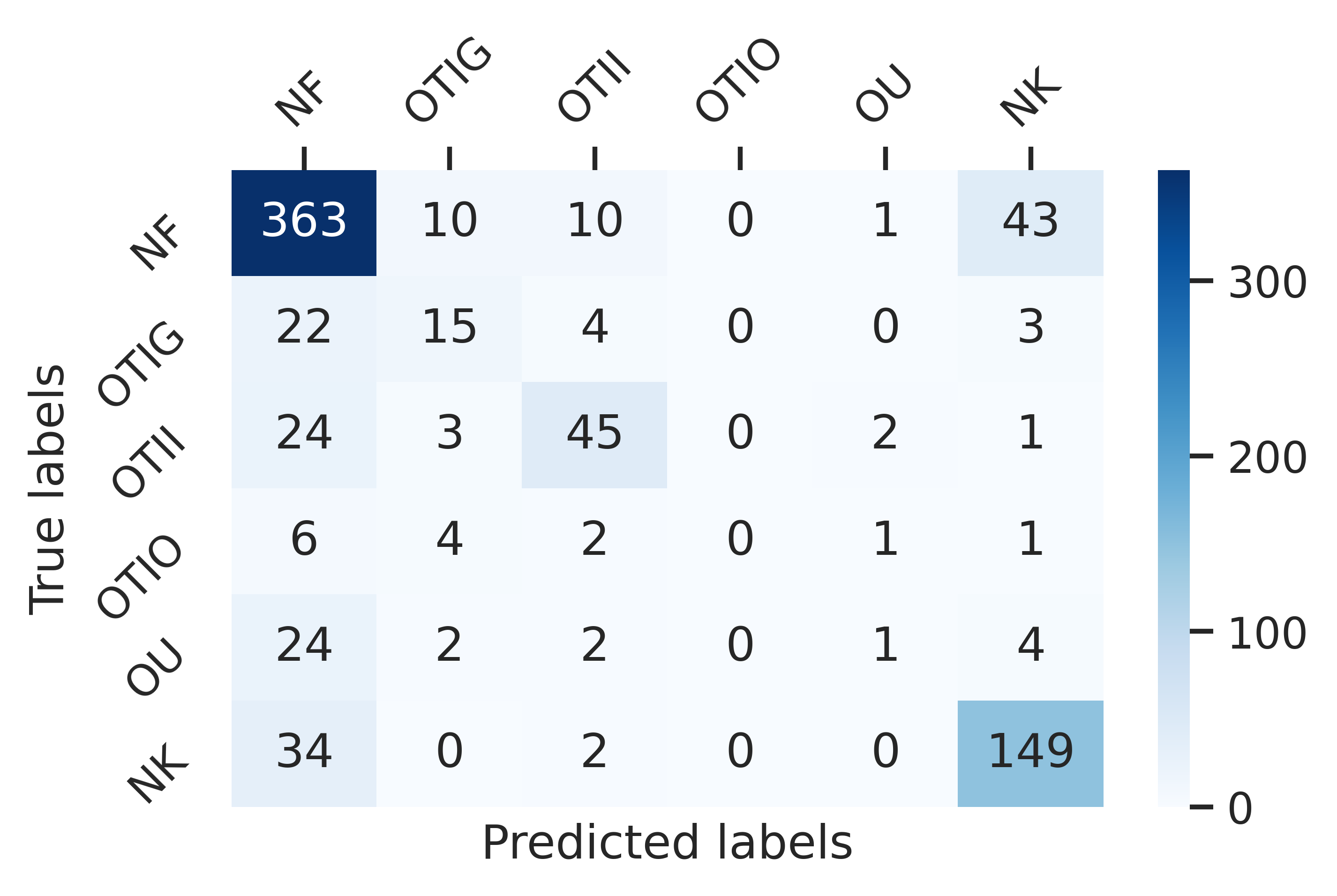}}
   
    \end{multicols}
 \caption{Confusion matrix of the best models for each language task}
 \label{confusion}
\end{figure*}

Table \ref{result} showed that XLM-R is the best performing for Tamil and Malayalam languages model, whereas, for Kannada, m-BERT is the best model. To get more insights, we present a detail error analysis of these models. The error analysis is carried out by using the confusion matrix (Figure~\ref{confusion}). Only NF and NT classes have a high true-positive rate (TPR) while in other classes, texts have mostly misclassified as non-offensive (Figure~\ref{confusion} (a)). However, among all six classes, OTIO got $0\%$ TPR where among $77$ texts more than $50\%$ wrongly classified as NF. The high-class imbalance situation might be the reason for this vulnerable outcome. Figure~\ref{confusion} (b) reveals that the model is biased towards NF class. Due to the insufficient number of instances, the model failed to identify none of the 23 OTIG texts. Among $27$ OTII texts, only $2$ have correctly classified. Meanwhile, we observe that only NF, OTII and NK classes have a high TPR where most of the cases, the model misclassified as NF class (Figure~\ref{confusion} (c)). In contrast, OTIO and OU classes have shown misclassification rate 100\% and 97\%.         

\section{Conclusion}
In this work, we have described and analyzed the system’s performance implemented as a participation in the offensive language identification shared task at EACL-2021. Initially, SVM, LR, LSTM, LSTM+Attention models have employed with tf-idf and word embedding features. Results indicate that ML ensemble achieved higher accuracy than DL methods. However, the outcomes are not promising for the available datasets. Code-mixing of multilingual texts might be a reason behind this. We applied transformer-based models to overcome this situation, which provides an astonishing rise in accuracy than ML and DL-based methods. Weighted $f_1$ score increased from $0.73$ to $0.76$, $0.88$ to $0.93$ and $0.48$ to $0.71$ for Tamil, Malayalam and Kannada language respectively. In future, the idea of ensemble technique could be adopted on transformer-based models to investigate the system’s overall performance.

\bibliography{anthology,eacl2021}

\begin{thebibliography}{23}
\expandafter\ifx\csname natexlab\endcsname\relax\def\natexlab#1{#1}\fi

\bibitem[{Akiwowo et~al.(2020)Akiwowo, Vidgen, Prabhakaran, and
  Waseem}]{alw-2020-online}
Seyi Akiwowo, Bertie Vidgen, Vinodkumar Prabhakaran, and Zeerak Waseem,
  editors. 2020.
\newblock \href {https://www.aclweb.org/anthology/2020.alw-1.0}
  {\emph{Proceedings of the Fourth Workshop on Online Abuse and Harms}}.
  Association for Computational Linguistics, Online.

\bibitem[{Aroyehun and Gelbukh(2018)}]{aroyehun2018aggression}
Segun~Taofeek Aroyehun and Alexander Gelbukh. 2018.
\newblock \href {https://www.aclweb.org/anthology/W18-4411} {Aggression
  detection in social media: Using deep neural networks, data augmentation, and
  pseudo labeling}.
\newblock In \emph{Proceedings of the First Workshop on Trolling, Aggression
  and Cyberbullying ({TRAC}-2018)}, pages 90--97, Santa Fe, New Mexico, USA.
  Association for Computational Linguistics.

\bibitem[{Bonanno and Hymel(2013)}]{bonanno2013cyber}
Rina~A Bonanno and Shelley Hymel. 2013.
\newblock \href {https://doi.org/10.1007/s10964-013-9937-1} {Cyber bullying and
  internalizing difficulties: Above and beyond the impact of traditional forms
  of bullying}.
\newblock \emph{Journal of youth and adolescence}, 42(5):685--697.

\bibitem[{Chakravarthi et~al.(2020)Chakravarthi, Muralidaran, Priyadharshini,
  and McCrae}]{chakravarthi2020corpus}
Bharathi~Raja Chakravarthi, Vigneshwaran Muralidaran, Ruba Priyadharshini, and
  John~Philip McCrae. 2020.
\newblock \href {https://www.aclweb.org/anthology/2020.sltu-1.28} {Corpus
  creation for sentiment analysis in code-mixed {T}amil-{E}nglish text}.
\newblock In \emph{Proceedings of the 1st Joint Workshop on Spoken Language
  Technologies for Under-resourced languages (SLTU) and Collaboration and
  Computing for Under-Resourced Languages (CCURL)}, pages 202--210, Marseille,
  France. European Language Resources association.

\bibitem[{Chakravarthi et~al.(2021)Chakravarthi, Priyadharshini, Jose, M,
  Mandl, Kumaresan, Ponnusamy, V, Sherly, and McCrae}]{dravidianoffensive-eacl}
Bharathi~Raja Chakravarthi, Ruba Priyadharshini, Navya Jose, Anand~Kumar M,
  Thomas Mandl, Prasanna~Kumar Kumaresan, Rahul Ponnusamy, Hariharan V,
  Elizabeth Sherly, and John~Philip McCrae. 2021.
\newblock Findings of the shared task on {O}ffensive {L}anguage
  {I}dentification in {T}amil, {M}alayalam, and {K}annada.
\newblock In \emph{Proceedings of the First Workshop on Speech and Language
  Technologies for Dravidian Languages}. Association for Computational
  Linguistics.

\bibitem[{Conneau et~al.(2020)Conneau, Khandelwal, Goyal, Chaudhary, Wenzek,
  Guzm{\'a}n, Grave, Ott, Zettlemoyer, and Stoyanov}]{conneau2019unsupervised}
Alexis Conneau, Kartikay Khandelwal, Naman Goyal, Vishrav Chaudhary, Guillaume
  Wenzek, Francisco Guzm{\'a}n, Edouard Grave, Myle Ott, Luke Zettlemoyer, and
  Veselin Stoyanov. 2020.
\newblock \href {https://doi.org/10.18653/v1/2020.acl-main.747} {Unsupervised
  cross-lingual representation learning at scale}.
\newblock In \emph{Proceedings of the 58th Annual Meeting of the Association
  for Computational Linguistics}, pages 8440--8451, Online. Association for
  Computational Linguistics.

\bibitem[{Dadvar et~al.(2013)Dadvar, Trieschnigg, Ordelman, and
  de~Jong}]{dadvar2013improving}
Maral Dadvar, Dolf Trieschnigg, Roeland Ordelman, and Franciska de~Jong. 2013.
\newblock \href {https://doi.org/10.1007/978-3-642-36973-5_62} {Improving
  cyberbullying detection with user context}.
\newblock ECIR'13, page 693–696, Berlin, Heidelberg. Springer-Verlag.

\bibitem[{Devlin et~al.(2019)Devlin, Chang, Lee, and
  Toutanova}]{devlin2019bert}
Jacob Devlin, Ming-Wei Chang, Kenton Lee, and Kristina Toutanova. 2019.
\newblock \href {http://arxiv.org/abs/1810.04805} {Bert: Pre-training of deep
  bidirectional transformers for language understanding}.

\bibitem[{Grave et~al.(2018)Grave, Bojanowski, Gupta, Joulin, and
  Mikolov}]{grave2018learning}
Edouard Grave, Piotr Bojanowski, Prakhar Gupta, Armand Joulin, and Tomas
  Mikolov. 2018.
\newblock \href {https://www.aclweb.org/anthology/L18-1550} {Learning word
  vectors for 157 languages}.
\newblock In \emph{Proceedings of the Eleventh International Conference on
  Language Resources and Evaluation ({LREC} 2018)}, Miyazaki, Japan. European
  Language Resources Association (ELRA).

\bibitem[{Hande et~al.(2020)Hande, Priyadharshini, and
  Chakravarthi}]{hande2020kancmd}
Adeep Hande, Ruba Priyadharshini, and Bharathi~Raja Chakravarthi. 2020.
\newblock \href {https://www.aclweb.org/anthology/2020.peoples-1.6}
  {{K}an{CMD}: {K}annada {C}ode{M}ixed dataset for sentiment analysis and
  offensive language detection}.
\newblock In \emph{Proceedings of the Third Workshop on Computational Modeling
  of People's Opinions, Personality, and Emotion's in Social Media}, pages
  54--63, Barcelona, Spain (Online). Association for Computational Linguistics.

\bibitem[{Kakwani et~al.(2020)Kakwani, Kunchukuttan, Golla, N.C.,
  Bhattacharyya, Khapra, and Kumar}]{kakwani2020inlpsuite}
Divyanshu Kakwani, Anoop Kunchukuttan, Satish Golla, Gokul N.C., Avik
  Bhattacharyya, Mitesh~M. Khapra, and Pratyush Kumar. 2020.
\newblock \href {https://doi.org/10.18653/v1/2020.findings-emnlp.445}
  {{I}ndic{NLPS}uite: Monolingual corpora, evaluation benchmarks and
  pre-trained multilingual language models for {I}ndian languages}.
\newblock In \emph{Findings of the Association for Computational Linguistics:
  EMNLP 2020}, pages 4948--4961, Online. Association for Computational
  Linguistics.

\bibitem[{Kumar et~al.(2020)Kumar, Ojha, Lahiri, Zampieri, Malmasi, Murdock,
  and Kadar}]{trac-2020-trolling}
Ritesh Kumar, Atul~Kr. Ojha, Bornini Lahiri, Marcos Zampieri, Shervin Malmasi,
  Vanessa Murdock, and Daniel Kadar, editors. 2020.
\newblock \href {https://www.aclweb.org/anthology/2020.trac-1.0}
  {\emph{Proceedings of the Second Workshop on Trolling, Aggression and
  Cyberbullying}}. European Language Resources Association (ELRA), Marseille,
  France.

\bibitem[{Maiya(2020)}]{maiya2020ktrain}
Arun~S. Maiya. 2020.
\newblock \href {http://arxiv.org/abs/2004.10703} {ktrain: A low-code library
  for augmented machine learning}.

\bibitem[{Mandl et~al.(2019)Mandl, Modha, Majumder, Patel, Dave, Mandlia, and
  Patel}]{mandl2019overview}
Thomas Mandl, Sandip Modha, Prasenjit Majumder, Daksh Patel, Mohana Dave,
  Chintak Mandlia, and Aditya Patel. 2019.
\newblock \href {https://doi.org/10.1145/3368567.3368584} {Overview of the
  hasoc track at fire 2019: Hate speech and offensive content identification in
  indo-european languages}.
\newblock In \emph{Proceedings of the 11th Forum for Information Retrieval
  Evaluation}, FIRE '19, page 14–17, New York, NY, USA. Association for
  Computing Machinery.

\bibitem[{Mikolov et~al.(2013)Mikolov, Sutskever, Chen, Corrado, and
  Dean}]{mikolov2013distributed}
Tomas Mikolov, Ilya Sutskever, Kai Chen, Greg Corrado, and Jeffrey Dean. 2013.
\newblock \href {http://arxiv.org/abs/1310.4546} {Distributed representations
  of words and phrases and their compositionality}.

\bibitem[{Mubarak et~al.(2017)Mubarak, Darwish, and
  Magdy}]{mubarak-etal-2017-abusive}
Hamdy Mubarak, Kareem Darwish, and Walid Magdy. 2017.
\newblock \href {https://doi.org/10.18653/v1/W17-3008} {Abusive language
  detection on {A}rabic social media}.
\newblock In \emph{Proceedings of the First Workshop on Abusive Language
  Online}, pages 52--56, Vancouver, BC, Canada. Association for Computational
  Linguistics.

\bibitem[{Ranasinghe and Zampieri(2020)}]{ranasinghe2020multilingual}
Tharindu Ranasinghe and Marcos Zampieri. 2020.
\newblock \href {https://doi.org/10.18653/v1/2020.emnlp-main.470} {Multilingual
  offensive language identification with cross-lingual embeddings}.
\newblock In \emph{Proceedings of the 2020 Conference on Empirical Methods in
  Natural Language Processing (EMNLP)}, pages 5838--5844, Online. Association
  for Computational Linguistics.

\bibitem[{Sharif et~al.(2020)Sharif, Hossain, and Hoque}]{sharif2020techtexc}
Omar Sharif, Eftekhar Hossain, and Mohammed~Moshiul Hoque. 2020.
\newblock \href {http://arxiv.org/abs/2012.11420} {Techtexc: Classification of
  technical texts using convolution and bidirectional long short term memory
  network}.

\bibitem[{Sharif et~al.(2021)Sharif, Hossain, and Hoque}]{sharif2021combating}
Omar Sharif, Eftekhar Hossain, and Mohammed~Moshiul Hoque. 2021.
\newblock \href {http://arxiv.org/abs/2101.03291} {Combating hostility:
  Covid-19 fake news and hostile post detection in social media}.

\bibitem[{Tokunaga and Makoto(1994)}]{tokunaga1994text}
Takenobu Tokunaga and Iwayama Makoto. 1994.
\newblock Text categorization based on weighted inverse document frequency.
\newblock In \emph{Special Interest Groups and Information Process Society of
  Japan (SIG-IPSJ}. Citeseer.

\bibitem[{Vaswani et~al.(2017)Vaswani, Shazeer, Parmar, Uszkoreit, Jones,
  Gomez, Kaiser, and Polosukhin}]{vaswani2017attention}
Ashish Vaswani, Noam Shazeer, Niki Parmar, Jakob Uszkoreit, Llion Jones,
  Aidan~N. Gomez, Lukasz Kaiser, and Illia Polosukhin. 2017.
\newblock \href {http://arxiv.org/abs/1706.03762} {Attention is all you need}.

\bibitem[{Zampieri et~al.(2019{\natexlab{a}})Zampieri, Malmasi, Nakov,
  Rosenthal, Farra, and Kumar}]{zampieri2019predicting}
Marcos Zampieri, Shervin Malmasi, Preslav Nakov, Sara Rosenthal, Noura Farra,
  and Ritesh Kumar. 2019{\natexlab{a}}.
\newblock \href {https://doi.org/10.18653/v1/N19-1144} {Predicting the type and
  target of offensive posts in social media}.
\newblock In \emph{Proceedings of the 2019 Conference of the North {A}merican
  Chapter of the Association for Computational Linguistics: Human Language
  Technologies, Volume 1 (Long and Short Papers)}, pages 1415--1420,
  Minneapolis, Minnesota. Association for Computational Linguistics.

\bibitem[{Zampieri et~al.(2019{\natexlab{b}})Zampieri, Malmasi, Nakov,
  Rosenthal, Farra, and Kumar}]{zampieri2019semeval}
Marcos Zampieri, Shervin Malmasi, Preslav Nakov, Sara Rosenthal, Noura Farra,
  and Ritesh Kumar. 2019{\natexlab{b}}.
\newblock \href {https://doi.org/10.18653/v1/S19-2010} {{S}em{E}val-2019 task
  6: Identifying and categorizing offensive language in social media
  ({O}ffens{E}val)}.
\newblock In \emph{Proceedings of the 13th International Workshop on Semantic
  Evaluation}, pages 75--86, Minneapolis, Minnesota, USA. Association for
  Computational Linguistics.

\end{thebibliography}
\bibliographystyle{acl_natbib}

\end{document}